\documentclass[letterpaper, 10 pt, conference]{ieeeconf}  

\IEEEoverridecommandlockouts                              

\usepackage{amsmath} 
\usepackage{amssymb}  

\usepackage[pdftex]{graphicx}       
\usepackage[dvipsnames]{xcolor}
\usepackage{algorithm}
\usepackage[noend]{algpseudocode}
\usepackage{float}
\usepackage{siunitx}
\usepackage{hyperref}
\makeatletter
\let\NAT@parse\undefined
\makeatother
\usepackage[numbers]{natbib}

\newcommand{\figref}[1]{Fig.~\ref{#1}}

\newcommand{\myparagraph}[1]{\vspace{0.03in}\noindent\textbf{#1}}
\newcommand*{\Cdot}{\raisebox{-0.25ex}{\scalebox{1.75}{$\cdot$}}}

\newboolean{draft-mode}
\setboolean{draft-mode}{true}
\newcommand{\sidenote}[1]{\ifthenelse{\boolean{draft-mode}}{\marginpar{\tiny\raggedright\textsf{\hspace{0pt}#1}}}{}}
\DeclareRobustCommand{\arnote}[1]{\ifthenelse{\boolean{draft-mode}}{\textcolor{blue}{\textbf{AR: #1}}}{}}
\DeclareRobustCommand{\ncdnote}[1]{\ifthenelse{\boolean{draft-mode}}{\textcolor{green}{\textbf{NCD: #1}}}{}}

\title{\LARGE \bf Pneumatic Shape-shifting Fingers to Reorient and Grasp
}

\author{\authorblockN{Nikhil Chavan-Dafle, Kyubin Lee and Alberto Rodriguez}
\authorblockA{Massachusetts Institute of Technology\\
    {\tt\small \{nikhilcd, kyubinl, albertor\}@mit.edu}\vspace{8mm}}
\includegraphics[scale=0.99]{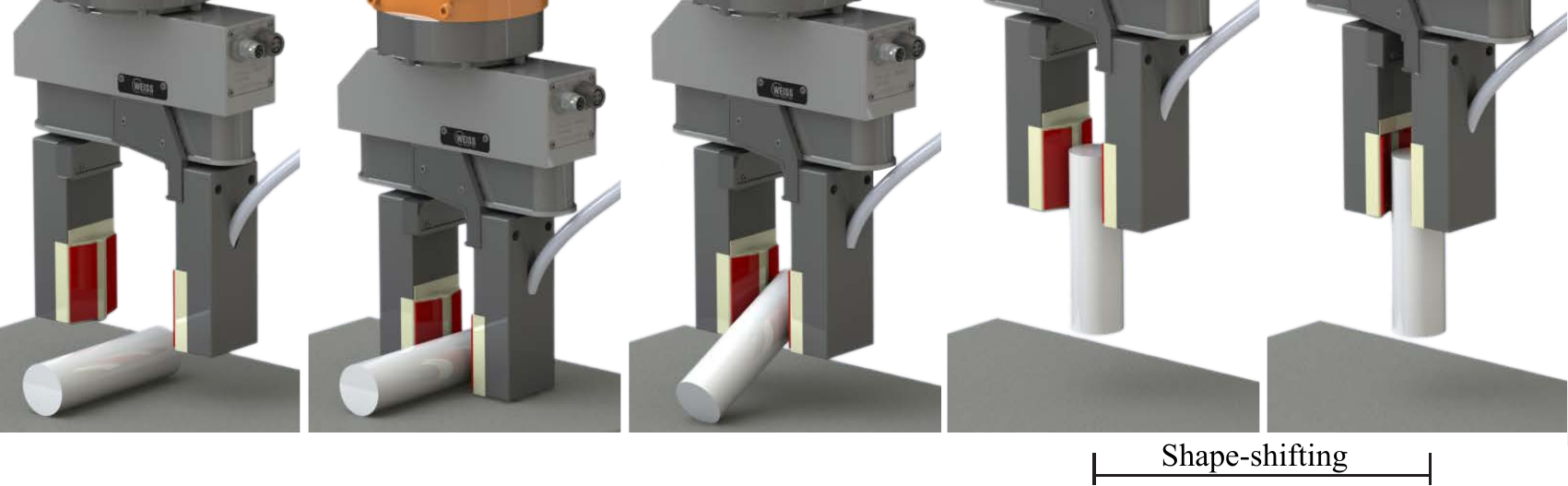}
\vspace{-3mm}}

\begin{document}

\maketitle
\thispagestyle{empty}
\pagestyle{empty}

\begin{abstract}

We present pneumatic shape-shifting fingers to enable a simple parallel-jaw gripper for different manipulation modalities. 
By changing the finger geometry, the gripper effectively changes the contact type between the fingers and an object to facilitate distinct manipulation primitives.
In this paper, we demonstrate the development and application of shape-shifting fingers to reorient and grasp cylindrical objects. 
The shape of the fingers changes based on the air pressure inside them and attains two distinct geometric forms at high and low pressure values. In our implementation, the finger shape switches between a wedge-shaped geometry and V-shaped geometry at high and low pressure, respectively.
Using the wedge-shaped geometry, the fingers provide a point contact on a cylindrical object to pivot it to a vertical pose under the effect of gravity. By changing to V-shaped geometry, the fingers localize the object in the vertical pose and securely hold it. 
Experimental results show that the smooth transition between the two contact types allows a robot with a simple gripper to reorient a cylindrical object lying horizontally on a ground and to grasp it in a vertical pose.
\end{abstract}


\section{Introduction}
\label{sec:intro}

A few part geometries and tools make up a large share of tasks in industrial assembly. Cylindrical objects, followed by prismatic ones, have been identified to as predominant parts in manufacturing industries~\cite{Groover86,Chen82}. Grippers play a crucial role of handling these objects. Bracken~\cite{Bracken86} notes that $98\%$ of these gripper are two-finger grippers, mostly in a parallel-jaw form. A parallel-jaw gripper often compromises the dexterity of part handling for the benefit of simplicity and robustness.
A common task then of grasping a cylindrical object lying horizontally on a ground in an upright configuration becomes a challenging problem. Industrial robots are often supported with part feeders which take care of the necessary reorientation. This approach leads to expensive solutions which are time consuming to setup and reconfigure.

We present pneumatic shape-shifting fingers that attach to a parallel-jaw gripper and enable it for specific manipulation modalities. The shape of the fingers changes as a function of air pressure in them and the gripping force. 
The key enabler is a shape-shifting membrane which is composed of rigid and flexible members. The rigid members provide a structure to the distinct geometries of the membrane at different pressure levels. The flexible members act like hinges to allow the transition between two different geometries. 
By changing the finger shape, the gripper actively changes the contact type between the object and the fingers that is appropriate for distinct manipulation modalities. 

In this paper, we discuss the development of the shape-shifting fingers for a task of reorienting and grasping cylindrical objects. The figure above shows the application of shape-shifting fingers to reconfigure a cylindrical object from a horizontal pose to a vertical one.
Initially, the object on the ground is picked up with low gripping force and with high air pressure inside the fingers. The fingers take the form of triangular prisms/wedges and make two opposite point contacts on the cylindrical object. 
As the object is lifted up, it pivots to the vertical orientation under gravity. The air pressure is then reduced and the gripping force is increased. The finger shape smoothly changes to that of a V-shaped cavity. The V-shaped cavity localizes the cylindrical object and grasps it securely.

We validate the functionality of the shape-shifting fingers on a manipulation platform with industrial robot arm and a parallel-jaw gripper with force control. Experimental results show that the smooth transition between the two contact geometries facilitated by the shape-shifting fingers enable the robot to  reorient a cylindrical object lying horizontally on a ground and to securely grasp it in a vertical pose.

With shape-shifting pneumatic fingers that actively change the form for different functionalities, we provide robots with simple grippers a robust capability to reorient objects and use them for the desired application.

\section{Related Work}
\label{sec:related}

Robotics community considers two parallel approaches to manipulation dexterity. One approach equips robots with multi-finger hands and a reasoning capability to control the fine motions at the fingers to do a large set of dexterous manipulations~\cite{salisbury1982ahf, Brock88, Trinkle1990, Rus1999a}. In contrast, the other approach focuses on a particular type of manipulations using a task-specific hardware and simple motion primitives. Industry has often preferred the latter approach for its simplicity and robustness.

Pivoting is one of the motion primitives studied extensively. Rao et al.~\cite{Rao96} explored the application of pivoting
for part reorientation for assembly automation. They use gravity to reorient grasped objects, as we do in
this paper and our previous work~\cite{ChavanDafle2014, ChavanDafle2015b}. Holladay et al.~\cite{holladay15} propose a framework for pivoting with dynamic motions of the robot arm. Vi\~{n}a B et al.~\cite{kragic_pivoting2} study the use of tactile and visual feedback to control the pivoting action. In our prior work, we explore the use of features in the environment to reorient, e.g., to pivot, an object in the grasp~\cite{ChavanDafle2015a,ChavanDafle2017,ChavanDafle2018a}. Recent work by Hou et al.~\cite{Hou2018} demonstrate the application of a custom designed gripper that can transition between pivoting grasp and firm grasp for planning wide variety of in-hand manipulations by sequences of pivoting and rolling. 

The work in this paper is motivated by our prior work on the two-phase gripper and addresses many of the limitations of the two-phase gripper in~\cite{ChavanDafle2015b}.
Our work in~\cite{ChavanDafle2015b} presents an approach to transition between two different contact geometries using an elastic strip. This approach requires adaptation of the stiffness of the elastic strip to manipulate objects of different weights even when the shape and size of the objects remains the same. Moreover, this method requires the robot to know the position of the object precisely and to place the gripper at an accurate location to pivot the object in the grasp. Though our prior work in~\cite{ChavanDafle2015b} and the work presented in this paper look at the same application for motivation, the finger design presented in this paper provide more comprehensive and robust solution to reorient and grasp cylindrical and prismatic objects. Unlike the design in~\cite{ChavanDafle2015b}, the same pneumatic shape-shifting fingers can be used for objects of different sizes and weights, as long as the objects are of the same geometry. 

The recent work on soft robotic hands explore the role of pneumatics to design novel gripper mechanisms. Deimel and Brock~\cite{Brock16} study the design and control of novel actuators to configure a multi-finger soft gripper. Rus and Tolley~\cite{Rus15nature} and Marchese et al.~\cite{Rus15} discuss the applications of novel material and manufacturing techniques for designing soft gripper with pneumatic control.  
Brown~\cite{Brown10} presented the idea of universal gripper which exploits the combination of conformable materials and negative air pressure to pick a large variety of objects. Developing a versatile gripper for picking objects has been the focus of most of these works.
In contrast, our work in this paper focuses on the use of pneumatics for changing the contact geometry to derive different manipulation modalities, for example, pivoting and grasping in this paper.

\section{Problem Motivation}
\label{sec:motivation}
\begin{figure}
\centering
 \includegraphics[scale=0.975]{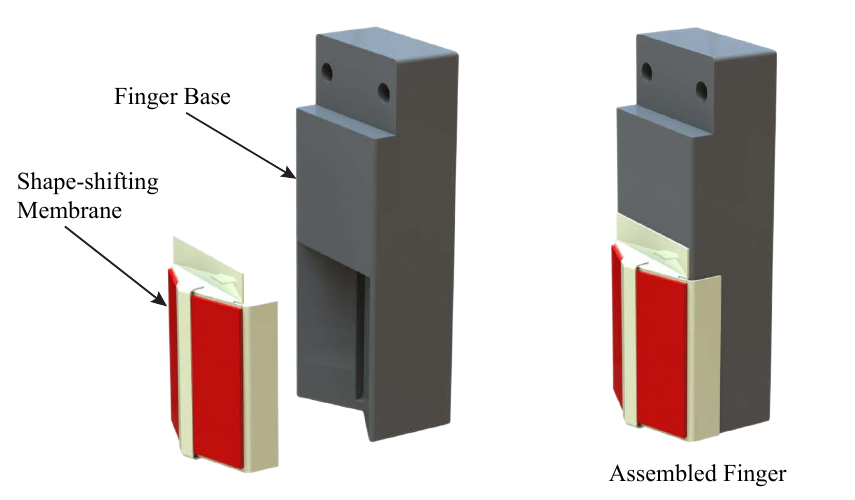}
\caption{(left) The key components in the design of the shape-shifting finger and (right) the assembled finger.}
    \label{fig:fing_des}
\end{figure} 
We present the idea of pneumatic fingers that change their geometry to derive different functionality.
For the scope of this paper, we focus on the application of such fingers for reorientating a cylindrical object to a vertical pose and then securely grasping it. This is one of the common tasks in an industrial assembly that robots need to do, for example,  when picking up objects from a table or a conveyor and then fitting them into a product in the upright pose.

To pivot an object in a grasp under the effect of gravity, there needs to be low torsional friction at the finger contacts. Contacts of small area provide small torsional friction. To securely hold an object in the grasp, the fingers need to provide specific kinematic constraints to localize the object and sufficient frictional resistance to prevent any motion of the object in the grasp.
For a gripper to be able to perform these two tasks in a sequence, it needs to be able to change the contact geometry between the fingers and the object accordingly.

To meet the necessary conditions for a gripper to be able to pivot and grasp an object and to address the limitations of our prior work in~\cite{ChavanDafle2015b}, we list the following functional requirements for the fingers:

\begin{enumerate}
    \item The fingers should provide contacts of small area at low grasping force to facilitate pivoting of the object in the grasp.
    
    \item The fingers should provide contacts with specific kinematic constraints and sufficient friction to securely grasp the object in the upright pose.
    
    \item The fingers, without any modifications, should be able to reorient and grasp objects of different sizes and weights, if the objects have the same shape.
    
    \item The functionality of the fingers should not be sensitive to the precise placement of the gripper over the object. 
\end{enumerate}

\section{Design of Pneumatic Shape-shifting Fingers}
\label{sec:design}
The functional requirements for the fingers guide different aspects of the design of the fingers. \figref{fig:fing_des} shows the design of the assembled shape-shifting finger and its components. In the following sections we explain the constraints that govern the choice of these components and highlight their role in meeting the functional requirements.

\subsection{Shape Shifting Membrane}
\label{sec:membrane}
\begin{figure}
\centering
 \includegraphics[scale=0.975]{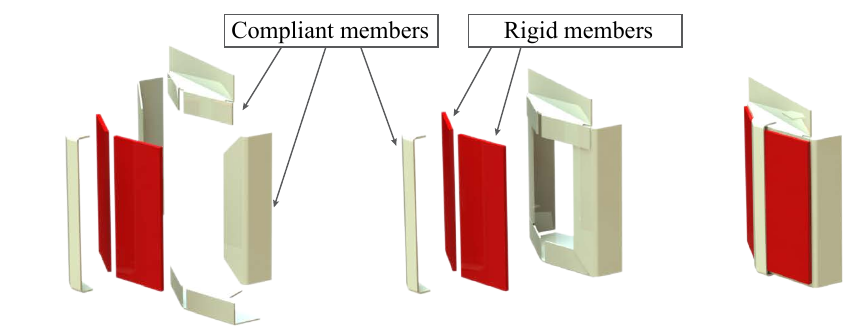}
\caption{The detailed components of the shape-shifting membrane. The rigid members define the structural shape of the membrane, while the  compliant parts act as hinges or fold lines to switch the structural shape of the membrane when desired. The rigid members are plastic panels while the compliant members are made up of rubber.}
    \label{fig:membrane_des}
    \vspace{2mm}
\end{figure} 
\begin{figure}
\centering
 \includegraphics[scale=0.95]{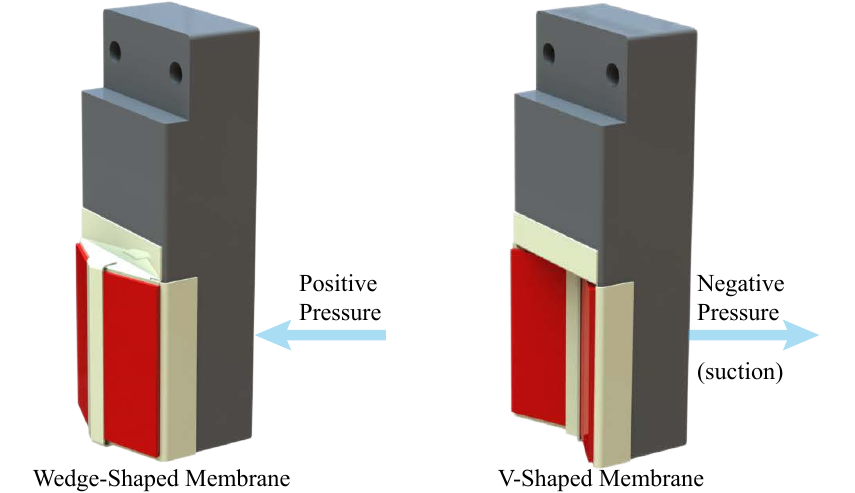}
\caption{The geometry of the shape shifting membrane changes when the air-pressure inside the finger is modulated.}
    \label{fig:shape_change}
\end{figure} 
The key component of the finger design is the membrane that takes particular shape at different air pressures. The membrane is made up of rigid and compliant members. The rigid members provide a structure to the geometry of the membrane at different air-pressure levels, while the compliant members act as hinges and allow the transition between these geometries. 
In our implementation, the rigid members of the shape-shifting membrane are hard plastic panels while the compliant members are cut from a rubber sheet as shown in \figref{fig:membrane_des}.

The tasks we want to perform using the gripper and the shape of the object govern the need for a distinct contact type and consequently the required geometry of the membrane of the finger. In this paper we focus on pivoting and grasping tasks for cylindrical objects. Given these task constraints, we select wedge-shaped and V-shaped membrane geometries. Using the wedge-shaped geometry, the fingers provide a point contact on a cylindrical object to pivot it to a vertical pose under the effect of gravity. By changing to V-shaped geometry, the fingers localize the object in the vertical pose and securely hold it. 
\begin{figure}
\centering
 \includegraphics[scale=0.95]{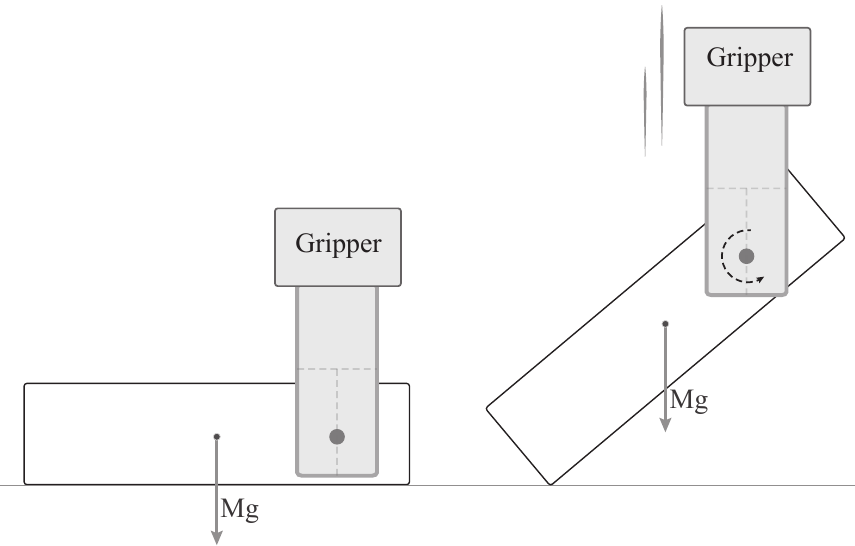}
\caption{The torque applied on the object as a result of gravitational force, pivots the object in the grasp. The contact between the cylindrical object and the wedge-shaped fingers remain a point contact during the pivoting phase. (This figure is adapted from a figure in~\cite{ChavanDafle2015b}.)}
    \label{fig:mechanics}
\end{figure} 
\begin{figure}
\centering
 \includegraphics[scale=0.95]{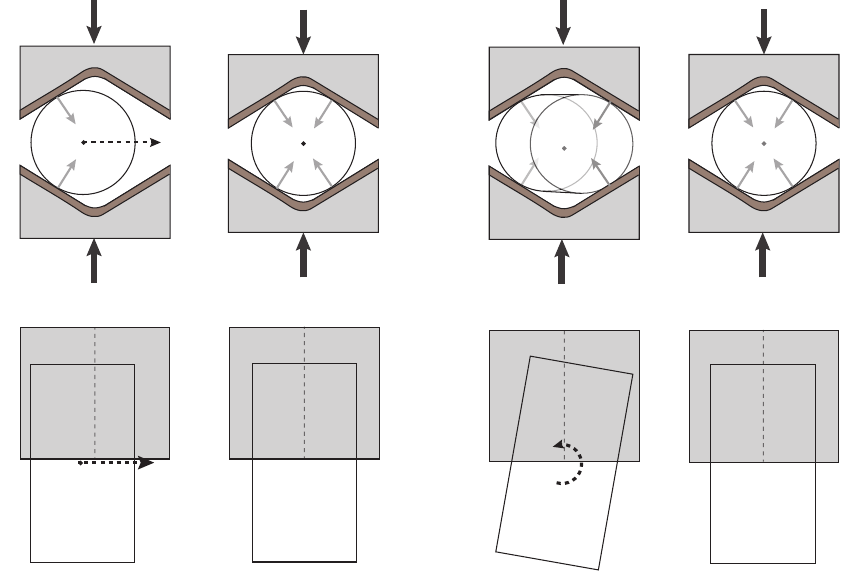}
\caption{The V-shaped membrane supported by rigid V-shaped cavity in the finger base aligns the object to the vertical pose by removing small deviations in the position or orientation of the object. (This figure is adapted from a figure in~\cite{ChavanDafle2015b}.)}
    \label{fig:vgroove}
\end{figure} 
%
\figref{fig:shape_change} shows the shape transformation of the shape-shifting membrane for two different  air pressures.

\myparagraph{Wedge-shaped membrane} provides a contact of small area (theoretically a point) on the object. Irrespective of small error in the position of a wedge shaped fingers, for cylindrical objects, the fingers will always make two apposite point contacts on the object. 
\figref{fig:mechanics} shows the sketch of object motion in the pivoting phase.
The contact with small area provide minimal torsional resistance to the rotation about the contact normal allowing the object to pivot in the grasp under the effect of gravity. 

\myparagraph{V-shaped membrane} provides kinematic constraints to localize the cylindrical object in the upright pose. If the object pose is not perfectly upright in the grasp after the pivoting phase, the V-groove corrects it. \figref{fig:vgroove} shows the application of the kinematic constraints from the V-grooves to localize the object.

\subsection{Finger Base}
\label{sec:base}
\begin{figure}
\centering
 \includegraphics[scale=0.975]{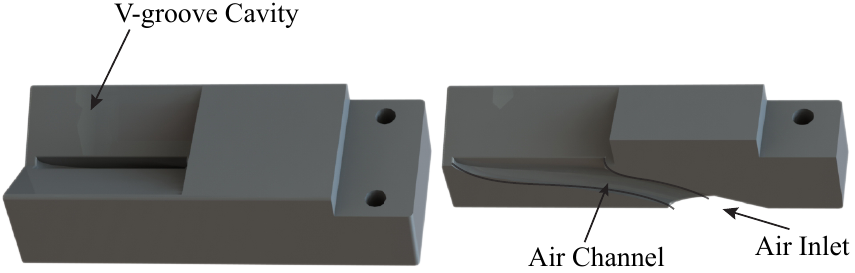}
\caption{The V-groove in the finger base provides a rigid support to the shape-shifting membrane during the grasping phase. (right) The air channel, shown in the section-view, supplies air to inflate/deflate the geometry defined by the membrane.}
    \label{fig:fing_base}
\end{figure} 
Base of the fingers holds the shape-shifting membrane and has an air channel to inflate or deflate the membrane to change its geometry.
\figref{fig:fing_base} shows the base of the finger and its section-view to show the internal  air channel.
The shape of the finger base is not critical because the shape of the membrane governs the contact between the object and the fingers. However, to provide rigid support to the membrane when it is deflated to form the V-shaped geometry for the grasping phase, we design the finger base with a V-groove cavity. The base of the finger is 3D printed.

\section{Setup and Operational Procedure}
\label{sec:setup}
\begin{figure}[t]
\centering
 \includegraphics[scale=0.47]{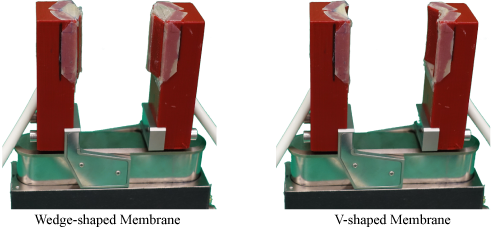}
\caption{Prototypes of shape-shifting fingers are mounted on a parallel-jaw gripper (Weiss Robotics WSG32). (left) Positive pressure inflates the membrane to a wedge-shaped geometry, while (right) the negative pressure deflates the membrane to create a V-groove geometry.}  \label{fig:exp_gripper}
\end{figure}

In this section we discuss the setup and operational procedure that enables a parallel-jaw gripper to use shape-shifting fingers to change their geometry and manipulate objects.
\figref{fig:exp_gripper} shows the prototype of shape-shifting fingers mounted on a parallel-jaw gripper.

\subsection{System Integration}
\label{sec:system}
\begin{figure}
\centering
 \includegraphics[scale=0.975]{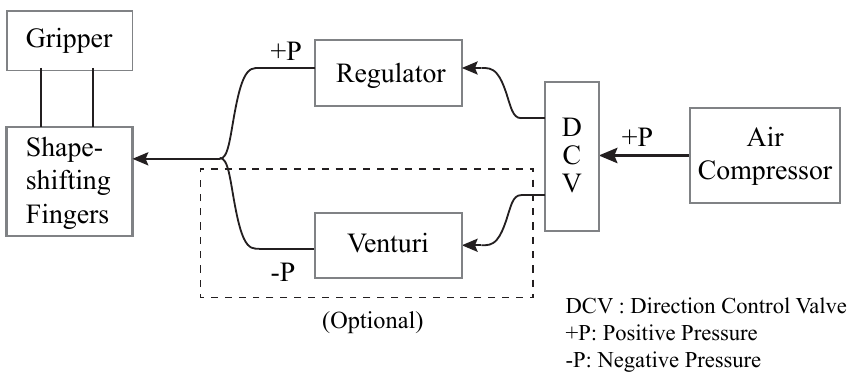}
\caption{The schematic of the system controlling the shape-shifting fingers mounted on a gripper. The negative pressure is not generally necessary as cutting off the air pressure and grasping the object with even a small force is often sufficient to trigger the shape-shift in the membrane.}
    \label{fig:circuit}
\end{figure}

The schematic of the setup that controls the shape of the fingers is shown in \figref{fig:circuit}. 
The air compressor supplies air at 20 psi pressure. The direction control valve routes the air through either a pressure regulator or to a venturi.

\myparagraph{Regulator} modulates the air pressure based on the pressure needed to inflate the shape-shifting membrane to a wedge-shaped geometry and to hold it in that form during the pivoting phase.

\myparagraph{Venturi} uses a compressed air to generate negative pressure down the line to create suction. This negative pressure deflates the membrane to a V-shaped geometry.

The use of negative pressure is necessary only if the grasping force is not sufficient to collapse the membrane to the V-groove shape. In practice, we did not find the negative pressure necessary for the objects we tested.

Most of the industrial robots provide internal channels to route the air supply through the arm, so the entire pneumatic control system for the fingers can be placed away from the manipulation platform if necessary.

\subsection{Operational procedure}
\label{sec:procedure}

We assume that a cylindrical object is presented in a horizontal configuration and the position of the object is known to the robot. To reorient the object to a vertical pose and to grasp it, the robot follows the procedure below:

\begin{itemize}
    \item[$\Cdot$] Supply high pressure air to inflate the shape-shifting membrane and create the wedge-shaped finger.
    
   \item[$\Cdot$] Place the gripper over the object offset to the center of the object. 
    
    \item[$\Cdot$] Grasp the object with low gripping force
    
    \item[$\Cdot$] Lift the object and let it pivot under gravity
    
    \item[$\Cdot$] Lower the air pressure and increase the gripping force to deflate the membrane and to grasp the object into the V-shaped finger.
\end{itemize}

\begin{figure*}
\centering
 \includegraphics[scale=0.95]{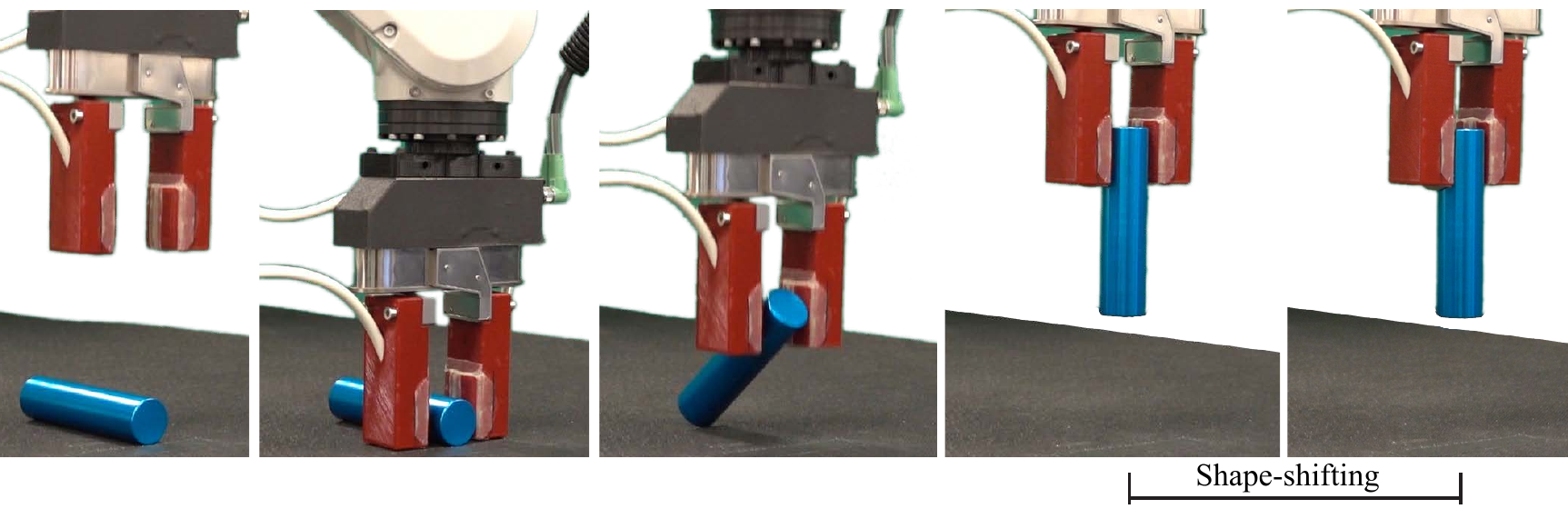}
\caption{Experimental run showing the steps involved in reorienting and grasping a cylindrical object using shape-shifting fingers.}
    \label{fig:2phase_exp_seq}
    \vspace{-2mm}
\end{figure*}

\figref{fig:2phase_exp_seq} shows the snapshots of an experimental run. It explains the operational procedure and also highlights the changes in the finger geometry during the process. Note the transition from the wedge-shaped finger to V-shaped finger at the end.

\section{Experimental Characterization}
\label{sec:characterization}

The geometry of the shape-shifting fingers is a function of the air pressure inside the fingers and the gripping force.
In this section, we experimentally characterize the gripping force required to pivot two different objects listed in Table \ref{tab:objects}. We record the air pressure required to maintain the wedge shape of the fingers for different gripping forces during the pivoting phase.
We can try to estimate these parameters and their relationship analytically. However, we find that characterizing these relationships empirically is more reliable and convenient, due to the robustness and repeatability of the functionality of the fingers.
\begin{table}[b]
\vspace{-5mm}
  \caption{Physical properties of the objects used.}
  \label{tab:objects}
	\centering
	\begin{tabular}{|l|l|c|r|}
         \hline
          \textbf{Object ID} & \textbf{Material} & \textbf{Dim [L, Diameter]} (mm) & \textbf{Mass} (g) \\ \hline
          Object 1  & Al 6061 & 100, 25 & 134\\ \hline  
          Object 2  & Polysulfone & 100, 25 & 67\\ \hline
    \end{tabular}
\end{table}
%
\begin{figure}
\centering
 \includegraphics[scale=1.05]{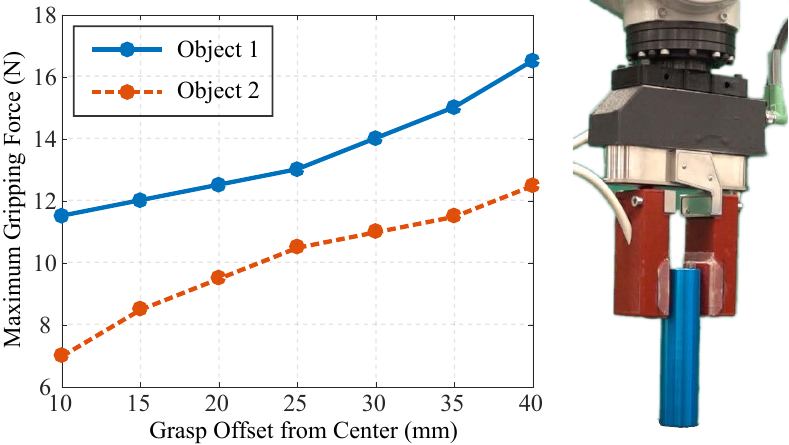}
\caption{The gripping force necessary to maintain a grasp on the object while allowing it to pivot under gravitational force.}
    \label{fig:offset_plot}
    \vspace{-3mm}
\end{figure} 

This characterization of the fingers is particularly important to get the expected performance when manipulating cylindrical objects of different sizes and weights and in different use conditions. If sufficient gripping force is not applied, the object may slip out of the grasp during pivoting. On the other hand, if more than necessary force is applied, the object may not pivot to the upright pose.
Similarly, if the pressure inside the fingers is not sufficient, the fingers will not maintain the wedge-shaped geometry. This will affect the contact type between the fingers and the object and the object may not pivot correctly to the vertical pose.

For the scope of this paper, we assume that the position of the object is provided to the robot. We manually reset the object to the same location before every experimental run.
 
\subsection{Grasp offset - Grip force Relationship}

For these experiments, the robot grasps the object at a location offset to the center and lifts the object to let it pivot under the gravitational force. The robot repeats the procedure for different offset values and gripping forces. We count a run as a successful run if the object pivots to a vertical pose as shows in \figref{fig:offset_plot}.

The minimum necessary gripping force is governed by the weight of the object. For the objects in Table~\ref{tab:objects}, the minimum gripping force is less than $5$ N, which is less than the force limit of the gripper we use. We focus only on characterizing the maximum gripping force for which the gripper can pivot the object. Any force between $5$N and the maximum bound on the gripping force will be able to successfully pivot the object in the grasp.

\figref{fig:offset_plot} shows the relationship between the offset distance and the maximum possible gripping force for a successful pivoting operation. 
As the offset distance is increased, the torque applied at the fingertips because of the weight of the object increases, so the object pivots to the vertical pose for a larger set of the gripping forces.
In practice, grasping the object at a larger offset is advantageous, because it provides a wider range for a valid gripping force, consequently providing robustness against uncertainty in the gripping force.

\subsection{Grip force - Air pressure Relationship}

The contact geometry plays an important role in governing the motion at the contacts. 
The geometry of the shape-shifting fingers is dependent on the the air pressure inside the fingers and the gripping force. To maintain the wedge-shaped geometry of the fingers for successful pivoting for different gripping forces, air pressure inside the fingers need to be varied accordingly.

If the air pressure inside the fingers is higher than the ideal pressure, it does not affect the functionality of the gripper adversely as the wedge-shaped geometry will still be maintained. However, if the air pressure is lower, the geometry is not maintained; the shape-shifting membrane becomes flat, providing large contact area, and consequently affecting the pivoting functionality of the gripper. We characterize the minimum pressure required for a successful pivoting phase for different gripping forces. We carried out multiple runs of pivoting phase by changing the gripping force and air pressure in every run. We recorded the pressure threshold below which the object no longer pivots to the vertical pose because the wedge-shaped geometry of the membrane is no longer well-maintained.
\begin{figure}
\centering
 \includegraphics[scale=0.975]{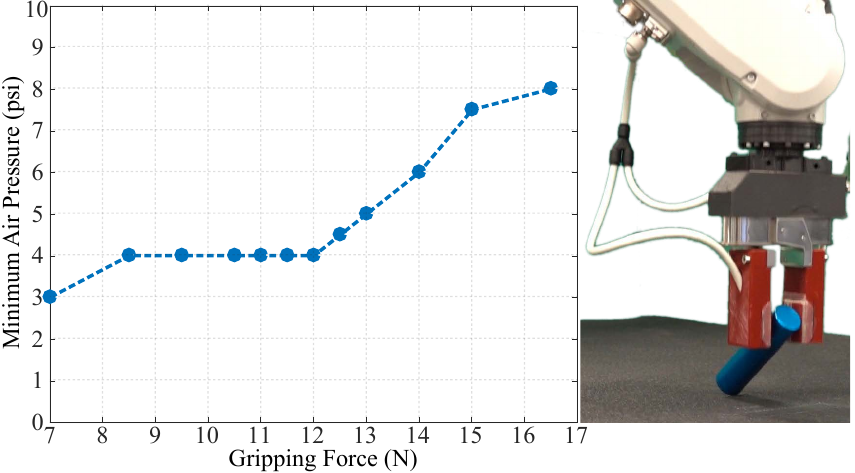}
\caption{The air pressure required to maintain the wedge-shaped geometry of the the membrane for different gripping forces to assure successful pivoting of the object in the grasp.}
    \label{fig:pressure_plot}
    \vspace{-3mm}
\end{figure} 


\figref{fig:pressure_plot} shows the minimum pressure required to maintain the wedge shape of the object for the pivoting phase. The trend observed in \figref{fig:pressure_plot} is intuitive from the force balance relationship. To resist the higher external force (gripping force), the pressure inside the finger geometry generated by the membrane needs to be increased. We observe minimal difference in the minimum air pressure for the cases with low gripping force. This could be because the pressure regulator we use can measure and control pressure with a resolution of $0.5$ psi. 
Any air pressure higher than this minimum pressure bound can work. Therefore, we do not need to have very precise identification of this pressure threshold.

The minimum air pressure values recorded in \figref{fig:pressure_plot} are valid for any object when using this finger design. However, the relationship between the grasp offset and the maximum grip force depends on the weight of the object, so needs to be reevaluated for new objects. The plots in \figref{fig:offset_plot} can be used for reference when doing the experiments for characterizing this relationship for the new objects.

\section{Discussion}
\label{sec:discussion}

We present an idea of pneumatic shape-shifting fingers that allow a simple parallel-jaw gripper to reorient and grasp objects.
In this paper, we demonstrate an implementation for the task of pivoting a cylindrical object to a vertical pose and then to securely grasp it.
The key enabler of this functionality is a shape-shifting membrane, composed of rigid and compliant members. This membrane takes the form of specific geometries at different air pressures. By choosing the geometry of the fingers for particular tasks, pivoting and grasping in this paper, we allow the gripper to provide necessary contact interaction with the parts and perform the desired tasks reliably. 

The design of shape-shifting finger is universal across objects of different sizes and weights, but of the same shape. The air pressure and the gripping force control the geometry of the finger and the contact interaction. By adjusting these two parameters, objects of different weights can be manipulated with no need of mechanical change in the finger design.

As a part of future work, we are interested to explore the application of shape-shifting fingers to more complex applications where we may need to change the finger geometry to multiple different geometries. An algorithmic approach for designing a shape-shifting membrane for new objects and new tasks opens up an exciting research direction.

Simple, easy-to-configure, and robust shape-shifting fingers extend the capabilities of a parallel jaw gripper for dexterous manipulation and flexible solutions that the new generation of industrial automation can benefit from.


\bibliographystyle{IEEEtranN} 
{\small \bibliography{ncd-case18}}

\end{document}